\documentclass[10pt]{article}
\usepackage[preprint]{spconf}
\usepackage[utf8]{inputenc} 
\usepackage[T1]{fontenc}    
\usepackage{hyperref}       
\usepackage{url}            
\usepackage{booktabs}       
\usepackage{amsfonts,amsmath,amssymb}       
\usepackage{nicefrac}       
\usepackage{microtype}      
\usepackage{lipsum}		
\usepackage{graphicx}
\usepackage{xcolor}

\usepackage{cite}
\usepackage{doi}
\usepackage{algorithm}
\usepackage{algpseudocode}
\usepackage{multirow}
\usepackage[super]{nth}
\usepackage{multirow}
\usepackage[normalem]{ulem}
\usepackage{enumitem}

\DeclareMathOperator*{\argmin}{arg\,min}
\renewcommand{\b}[1]{\textbf{#1}}

\title{StyleGAN-induced data-driven regularization for inverse problems}
\name{Arthur Conmy, Subhadip Mukherjee, and Carola-Bibiane Sch\"onlieb
\thanks{Corresponding author: Arthur Conmy. Code will be made available on request after successful publication.}}
\address{Department of Applied Mathematics and Theoretical Physics, University of Cambridge, UK \\
Emails: \{\texttt{asc70}, \texttt{sm2467}, \texttt{cbs31}\} \texttt{@cam.ac.uk}}

\begin{document}
\maketitle
\begin{abstract}
Recent advances in generative adversarial networks (GANs) have opened up the possibility of generating high-resolution photo-realistic images that were impossible to produce previously. The ability of GANs to sample from high-dimensional distributions has naturally motivated researchers to leverage their power for modeling the image prior in inverse problems. We extend this line of research by developing a Bayesian image reconstruction framework that utilizes the full potential of a pre-trained StyleGAN2 generator, which is the currently dominant GAN architecture, for constructing the prior distribution on the underlying image. Our proposed approach, which we refer to as \textit{learned Bayesian reconstruction with generative models} (L-BRGM), entails joint optimization over the \textit{style-code} and the input latent code, and enhances the expressive power of a pre-trained StyleGAN2 generator by allowing the style-codes to be different for different generator layers. Considering the inverse problems of image inpainting and super-resolution, we demonstrate that the proposed approach is competitive with, and sometimes superior to, state-of-the-art GAN-based image reconstruction methods.       
\end{abstract}
\section{Introduction}
Ill-posed inverse problems are encountered routinely in various imaging applications, wherein one seeks to estimate an unknown image $x\in \mathbb{R}^n$ from its degraded and noisy measurement $y\in\mathbb{R}^m$. Variational reconstruction circumvents ill-posedness by incorporating prior knowledge about $x$ via a regularizer $\psi:\mathbb{R}^n\rightarrow \mathbb{R}$, and computes an estimate $\hat{x}$ as 
\begin{equation}
    \hat{x}\in\underset{x}{\argmin}\,L\left(y,Ax\right)+\lambda\,\psi(x),
    \label{var_recon}
\end{equation}
where $A$ denotes the (typically ill-conditioned) measurement operator and $L$ measures data-fidelity. The role of $\psi$ is to penalize undesirable solutions. Some popular choices for $\psi$ include the Tikhonov regularization, total-variation (TV), and more recently, sparsity-promoting regularizers that encourage the image to be sparse in a suitable (analytical or learned) basis \cite{scherzer2009variational}. Notably, for a Gibbs-type prior $p(x)\propto \exp\left(-\lambda\,\psi(x)\right)$, \eqref{var_recon} can be interpreted as the Bayesian maximum a-posteriori probability (MAP) estimate of the unknown image $x$.  

The success of deep generative modeling in recent years serves as a major inspiration for learning the regularizer based on a dataset of training images \cite{nett_paper,ar_nips,bora_gancs,red_schniter,rare_deep_prior}. Such data-driven regularizers have been shown to significantly outperform their hand-crafted variants on a wide array of imaging inverse problems. Within the realm of data-driven regularizers, a particularly promising approach has been to seek a reconstruction $\hat{x}$ that lies in the range of a pre-trained generator $G:\mathcal{Z}\rightarrow \mathbb{R}^n$, where $\mathcal{Z}$ is a latent space whose dimension is significantly smaller than $n$. This amounts to reformulating \eqref{var_recon} as  \begin{equation}
    \hat{x}=G(\hat{z}), \text{\,where\,}\hat{z}\in\underset{z}{\argmin}\,L\left(y,AG(z)\right).
    \label{gan_inv}
\end{equation}
Training such a generator $G$ generally tends to suffer from \textit{mode collapse}, in which case $\text{range}(G)$ might fail to represent the image manifold in its entirety and there could be target images lying outside $\text{range}(G)$. In such scenarios, solving \eqref{gan_inv} fails to recover the image of interest, and one needs to allow for some flexibility for the target image to lie close to $\text{range}(G)$ through soft constraints. Such methods typically require computing a latent code for a given image, which calls for \textit{inverting} $G$, and are therefore referred to as GAN-inversion \cite{GANInversion}. Methods for inverting a GAN can be learning- or optimization-based, and we adopt the latter in our work. In particular, we extend the recently proposed \textit{Bayesian reconstruction through generative models} (BRGM) approach \cite{BRGM} by introducing a data-driven modeling of the regularizer on the latent space. Since our approach employs learned networks for prior modeling, we refer to it as learned-BRGM (L-BRGM).         

\section{The proposed L-BRGM method}
To construct the latent prior, L-BRGM relies on a pre-trained StyleGAN2, which is a recent state-of-the-art GAN architecture developed by Karras et al. \cite{StyleGAN2}. StyleGAN2 is an improved variant of its predecessor StyleGAN \cite{StyleGAN1}, named after its architecture's inspiration from the style transfer literature. In StyleGAN, the input latent space $\mathcal{Z}$ is warped into an intermediate disentangled feature space $\mathcal{W}$ (also called the style space) via a eight-layer fully connected mapping network $M$. The StyleGAN generator $G$, consisting of adaptive instance normalization (AdaIN) and convolution blocks, subsequently produces photo-realistic images from these disentangled features. Randomness is introduced in the generated samples by feeding noise into the generator. StyleGAN2 enhances the sample quality by redesigning the generator architecture and by introducing a novel path-length regularizer (PLR). The PLR seeks to ensure that the generator approximately preserves length, i.e., a fixed change in $w\in\mathcal{W}$ causes a fixed magnitude change in the sampled image. The effect of PLR is to promote orthogonality in the Jacobian matrix $J_G(w)=\frac{\partial G(w)}{\partial w}$ at any $w$. As remarked in \cite{StyleGAN2}, the PLR makes it easier to invert $G$, motivating the usage of StyleGAN2 in our approach.      

\subsection{A Bayesian formulation of L-BRGM}
The proposed L-BRGM approach has two key features: (i) formulating the reconstruction problem as a joint optimization over the style-code $w$ and the input latent code $z$, and (ii) using the pre-trained mapping network $M:\mathcal{Z}\rightarrow \mathcal{W}$ to model the distribution of $w$. Both of them emerge naturally from the Bayesian framework explained in the following. 

Let $p(y|w,z)$ be the probability density function (p.d.f.) of the measurement $y$ conditioned on the style-code $w$ and the input noise $z$. Given $w$, the distribution of $y$ is independent of $z$, and is determined by the generator $G$ and the measurement operator $A$. Consequently, we have that $p(y|w,z)=p(y|w)$. The posterior distribution of the joint latent code $(w,z)$ conditioned on $y$ can be expressed using Bayes' rule:   
\begin{equation}
    p(z, w|y)=\frac{p(y|w,z)p(w,z)}{p(y)}=\frac{p(y|w)p(w,z)}{p(y)}.
    \label{bayes_derive1}
\end{equation}
Further, $p(w,z)$ factors as $p(w,z)=p(w|z)p(z)$, leading to 
\begin{equation*}
    \log p(z, w|y) = \log p(y|w) + \log p(w|z) + \log p(z) - \log p(y).
\label{mapping_net_MAP_loss}
\end{equation*}
Since $w$ is obtained by applying the mapping network $M$ on $z$, it is natural to model $p(w|z)$ as a Gaussian perturbation around $M(z)$, i.e., $p(w|z)=\mathcal{N}\left(w;M(z),\lambda_{\text{map}}^{-1}\,I\right)$. Subsequently, noting that $p(z)=\mathcal{N}\left(z;0,\lambda_{\text{lat}}^{-1}I\right)$, and assuming the same measurement noise model considered in \cite[eq(4)]{BRGM} (i.e., Gaussian noise in the pixel space as well as in the VGG-16 embedding space), the MAP estimate of $(w,z)$ reduces to 
\begin{align}
        \underset{(z,w)}{\min}\lambda_{\text{pix}}\|y-A\,G(w)\|_2^2 + \lambda_{\text{vgg}} \|\varphi(y)-\varphi(A\,G(w))\|_2^2 \nonumber\\ 
    +\lambda_{\text{map}}\|w-M(z)\|_2^2+\lambda_{\text{lat}}\|z\|_2^2,
\label{mapping_net_MAP_loss_simp1}
\end{align}
where $\varphi$ computes the VGG-16 features. Finally, we extend \eqref{mapping_net_MAP_loss_simp1} by optimizing over $w^{+}$, where $w^{+}=\left\{w_i\right\}_{i=1}^{L}$, allowing for different style-codes $w_i$ at different layers $i$ of the generator. The resulting optimization reads $\underset{(z,w^{+})}{\min} J(w^+,z)$, where 
\small
\begin{align}
        J(w^+,z)&:=\lambda_{\text{pix}}\|y-A\,G(w^{+})\|_2^2+\lambda_{\text{map}}\sum_{i=1}^{L}\|w_i-M(z)\|_2^2\nonumber\\
    &+ \lambda_{\text{vgg}} \|\varphi(y)-\varphi(A\,G(w^{+}))\|_2^2 
     + \lambda_{\text{lat}}\|z\|_2^2.
\label{formal_loss_term}
\end{align}
\normalsize
$J$ is minimized via an iterative optimization algorithm (namely, Adam \cite{kingma2017adam}) starting from an appropriate initialization (c.f. Sec. \ref{sec_mult_init}); and the final reconstructed image is computed as $\hat{x}=G(\hat{w}^+)$, where $\hat{w}^+$ is the optimal style-code.
\vspace{-0.3in}
\begin{figure*}[t]
    \centering
    \includegraphics[scale=0.23]{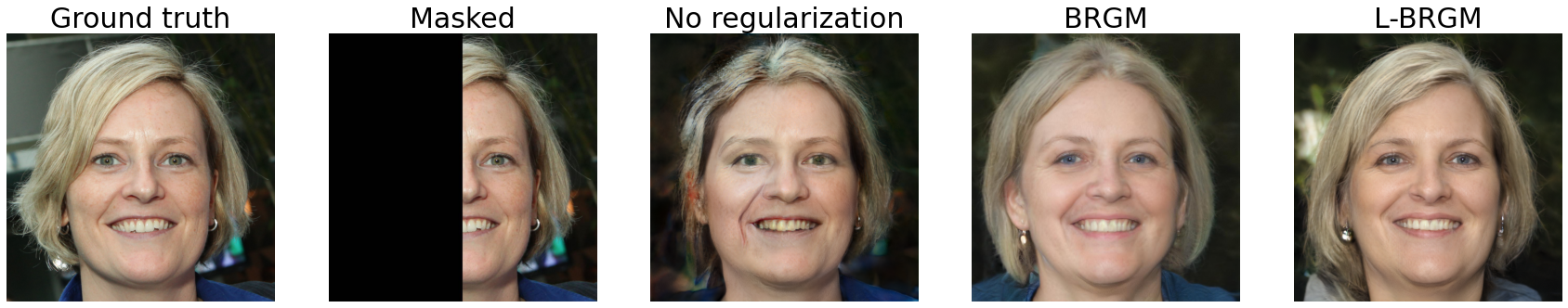}
    \caption{\small{Effect of regularization. Without any regularization, the reconstructed image tends to look unnatural}.} 
    \label{griddy}
\end{figure*}
\begin{center}
    \begin{figure*}[t]
        \centering
        \includegraphics[scale=0.2]{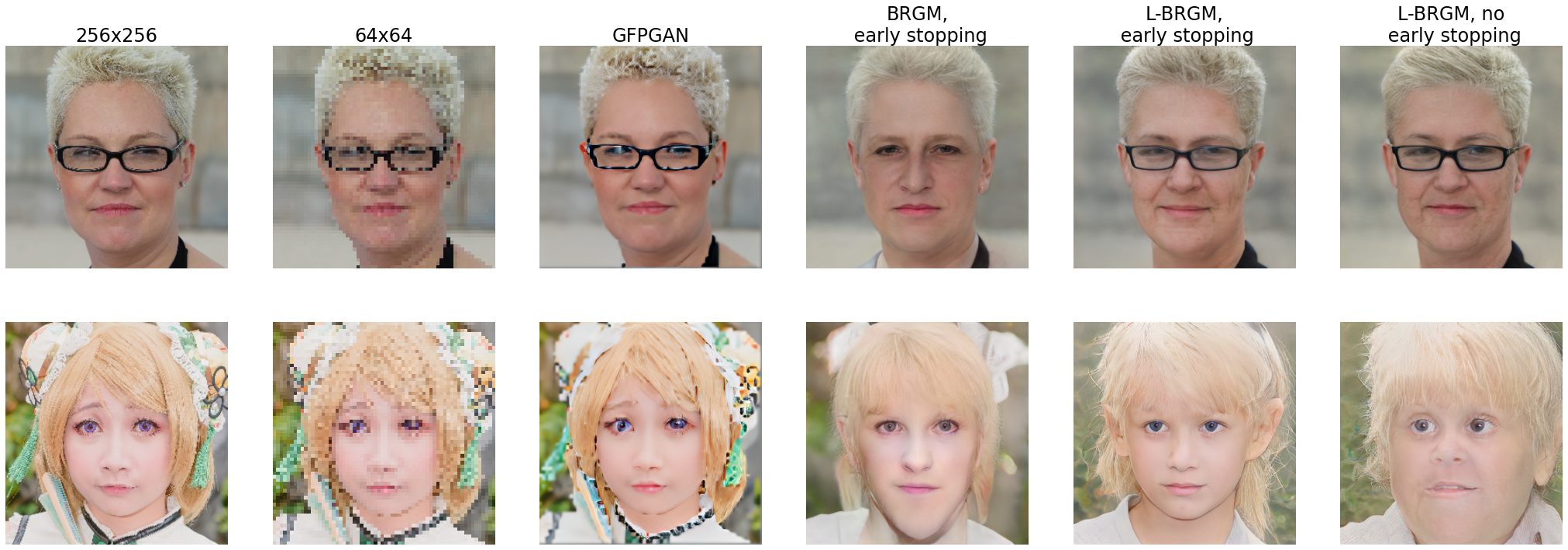}
        \caption{\small{Illustration of the effect of early stopping for L-BRGM and BRGM. Further, despite strong quantitative performance, we noted that GFP-GAN struggled to generate a natural-looking image}.}
        \label{methodcomp}
    \end{figure*}
\end{center}
\indent Since L-BRGM extends the BRGM approach by exploiting the StyleGAN2 mapping network for modeling $p(w)$, it is imperative to compare and contrast the Bayesian probability models for the two methods. As the ground-truth image $x$ can be modeled as $x=G(w)$ for an appropriate $w$, the BRGM approach seeks to maximize $p(x|y)\propto p(y|G(w))p(G(w))$. Subsequently, the image prior in BRGM is rewritten via the change of variable formula as $p(G(w))=p(w)\left(\text{det}\left(J_G(w)\right)\right)^{-1}$. Finally, making the simplifying assumption that the Jacobian $J_G(w)$ is not dependent on $w$ (i.e., $G(w)$ varies approximately linearly in $w$, which can be justified by the PLR during StyleGAN2 training), one ends up with the task of solving eq(6) posed in \cite{BRGM}. To model $p(w)$ in BRGM, one allows for the flexibility of having different $w_i$'s at different layers of $G$, while encouraging the $w_i$s to be similar via a cosine similarity prior. Additionally, each $w_i$ is modeled as a Gaussian random vector, with the parameters empirically estimated from a large number of samples fed through $M$.

L-BRGM extends this Bayesian formalism by formulating a Bayesian MAP estimation problem over the joint code $(w,z)$ and alleviates the need to hand-craft a suitable prior on $w$ by exploiting the mapping network $M$. L-BRGM also allows for different style-codes $w_i$ for $L=18$ different layers of $G$. Further, all $w_i$s are encouraged to be close to $M(z)$, thus ensuring mutual similarity among each pair of $w_i$s. L-BRGM does not require any Gaussian assumption on the style-codes $w_i$, but instead makes this assumption on the latent $z$, thereby making the framework more realistic (since $z$ is indeed sampled from a Gaussian distribution during training). 

\subsection{The effect of regularization}
\label{regularization_effects}
It is worth addressing the importance of modeling the distribution $p(w)$ of the style-code, which acts as a regularizer in the reconstruction process. If one optimizes the latent vector $w$ by solely minimizing the data-likelihood term (i.e., a combination of the pixel-wise error and VGG perceptual loss), the reconstructed images tend to have unnatural artifacts. This phenomenon is demonstrated in Fig. \ref{griddy} for inpainting on the FFHQ dataset consisting of human face images of resolution $1024^2$. The reconstructed image with no regularization on $w$ appears to be smooth around the inpainting boundary, but fails to recover important facial features (such as hair color and texture) far from this boundary. Notably, the L-BRGM reconstruction in Fig. \ref{griddy} matches the ground-truth more closely than BRGM. This highlights the fact that the regularization term on the style-code can lead to realizing the full capacity of the underlying generative model. While no regularization leads to worse reconstructions, hand-crafting the regularizer also limits the potential of the Bayesian framework considerably. 

\subsection{Key implementation details}
\label{sec_mult_init}
\label{sec_early_stopping}
The reconstruction loss $J(w^+, z)$ arising from the Bayesian MAP formalism is a non-convex objective of the joint latent code $(w^+, z)$. Therefore, it is generally challenging to devise an iterative algorithm that succeeds in recovering the global minimum regardless of the initialization. We employ the Adam algorithm developed by Kingma and Ba \cite{kingma2017adam} in view of its wide applicability for solving high-dimensional non-convex problems while requiring minimal hyper-parameter tuning. Nevertheless, the success of Adam in solving \eqref{formal_loss_term} depends to a great extent on careful initialization and the stopping rule, both of which are elucidated in the following.

\noindent \textit{Multiple initialization}: To initialize the latent code $z$ for \eqref{formal_loss_term}, we draw $N_1$ i.i.d. samples $\left\{z_{(i)}\right\}_{i=1}^{N_1}$ from the standard normal distribution, and select the best $z_{(i)}$ that minimizes the perceptual loss: $z_{\text{init}} = \underset{1\leq i\leq N_1}{\argmin}\,\|\varphi(y)-\varphi(A\,G(M(z_{(i)})))\|_2^2$. Computing $z_{\text{init}}$ does not require any expensive gradient computation and only entails forward passes through the generator and the mapping network. Such an initialization approach prevents any bias that might potentially arise from just one initial point and helps the reconstruction algorithm avoid local minima in the loss landscape. We found that $N_1=100$ sufficed for our experiments. 

\noindent \textit{Early stopping}: We found that for both BRGM and L-BRGM, the iterative scheme for reconstruction can potentially diverge if the iterations are not terminated appropriately. This phenomenon is illustrated in Fig. \ref{methodcomp} for the task of image super-resolution. The example in the second row of Fig. \ref{methodcomp} illustrates how early stopping might be necessary to prevent divergence of BRGM and L-BRGM from the image manifold, whereas the first row shows a representative example where such an early stopping may not be required. Therefore, we track the learned perceptual image patch similarity (LPIPS) metric \cite{LPIPS} between the ground-truth and the reconstructed image over the iterations and report the best LPIPS (i.e., the smallest) distance achieved during the iterative optimization process.
Empirically, we found that BRGM also attained the optimal LPIPS distance around the same iteration as L-BRGM. Such early stopping is not applicable to the GFP-GAN approach \cite{GFPGAN} considered for numerical comparison in Section \ref{sec_results_numerics}, since GFP-GAN is a learning-based inversion method, as opposed to BRGM and L-BRGM that employ optimization for inversion.
\begin{figure}[h]
    \centering
    \includegraphics[scale=0.135]{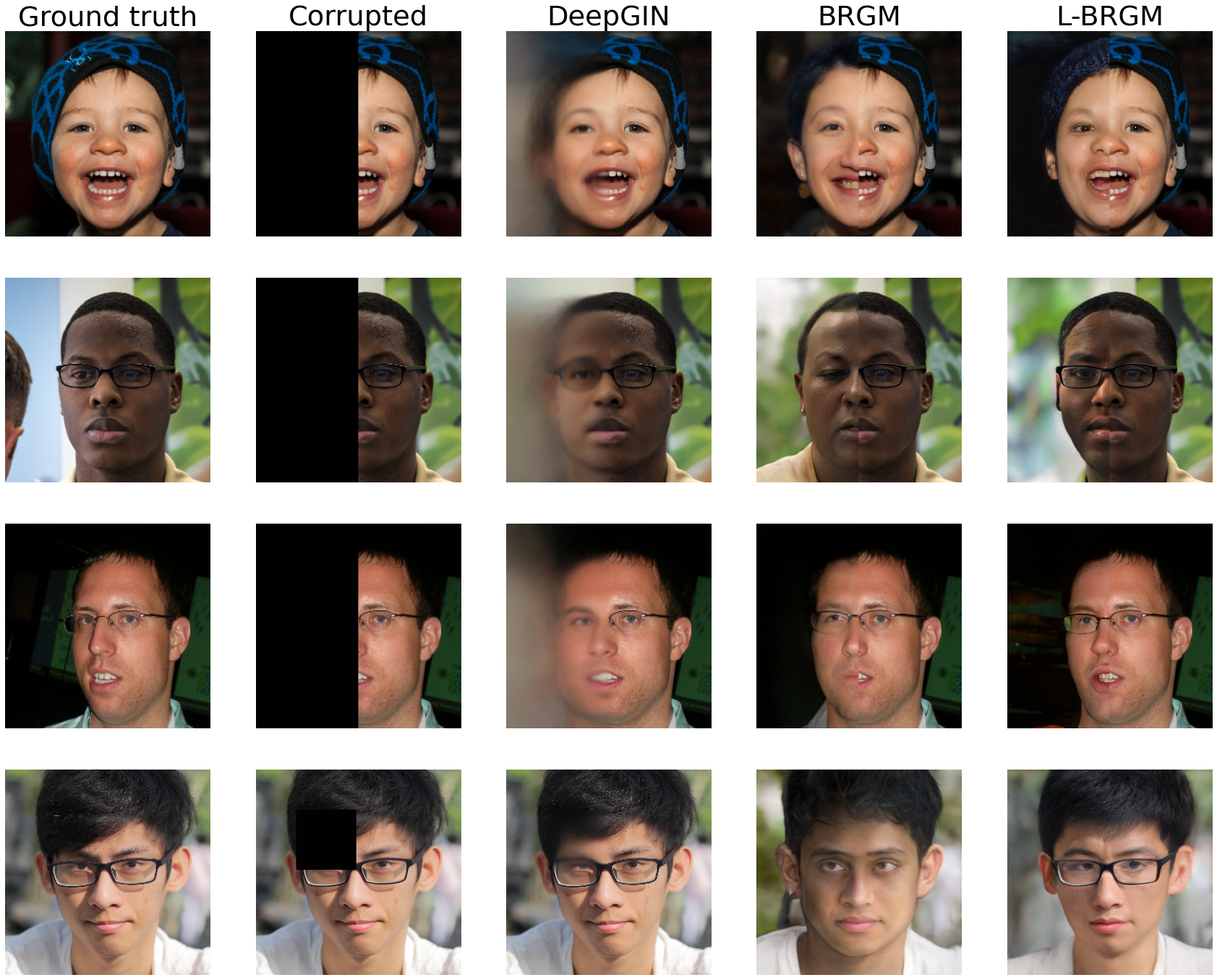}
    \caption{\small{Representative inpainting examples for the half mask (first three rows) and the eye-patch mask (last row). L-BRGM visibly outperforms BRGM and is competitive with DeepGIN.}}
    \label{halfmasking}
\end{figure}

\begin{table}[t]
        \resizebox{\columnwidth}{!}{%
        \begin{tabular}{llll|l|l|l|l|}
            \hline
            \multicolumn{1}{|l|}{\multirow{4}{*}{\rotatebox[origin=c]{90}{Inpainting}}} & \multicolumn{1}{l|}{\textbf{Input}} & \multicolumn{1}{l|}{\textbf{Mask}} & \textbf{Method}& \multicolumn{2}{c|}{\textbf{LPIPS}} & \multicolumn{2}{c|}{\textbf{SSIM}} \\ \cline{2-8} 
            \multicolumn{1}{|l|}{} & \multicolumn{1}{l|}{\multirow{3}{*}{$256^2$}} & \multicolumn{1}{l|}{\multirow{3}{*}{Half}} & DeepGIN & 1 & 0.303 & \b{59} & \b{0.625} \\ \cline{4-8}            
            \multicolumn{1}{|l|}{} & \multicolumn{1}{l|}{} & \multicolumn{1}{l|}{} & BRGM & 47 & 0.226 & 21 & 0.606 \\ \cline{4-8} 
            \multicolumn{1}{|l|}{} & \multicolumn{1}{l|}{} & \multicolumn{1}{l|}{} & L-BRGM & \b{51} & \b{0.224} & 19 & 0.594 \\ \cline{1-8}
            \multicolumn{1}{|l|}{\multirow{4}{*}{\rotatebox[origin=c]{90}{\small{Super-res.}}}} & \multicolumn{1}{l|}{\textbf{Input}} & \multicolumn{1}{l|}{\textbf{Output}} &\textbf{Method} & \multicolumn{2}{c|}{\textbf{LPIPS}} & \multicolumn{2}{c|}{\textbf{SSIM}} \\ \cline{2-8} 
            \multicolumn{1}{|l|}{} & \multicolumn{1}{l|}{\multirow{3}{*}{$64^2$ }} & \multicolumn{1}{l|}{\multirow{3}{*}{$256^2$}} & GFP-GAN & \b{72} & \b{0.231} & \b{89} & \b{0.534} \\ \cline{4-8} 
            \multicolumn{1}{|l|}{} & \multicolumn{1}{l|}{} & \multicolumn{1}{l|}{} & BRGM & 6 & 0.294 & 10 & 0.480 \\ \cline{4-8} 
            \multicolumn{1}{|l|}{} & \multicolumn{1}{l|}{} & \multicolumn{1}{l|}{} &  L-BRGM & 21 & 0.277 & 0 & 0.404 \\ \cline{1-8}
        
    
    \end{tabular}
    }
    \caption{\small{Comparison of L-BRGM with competing GAN-based approaches for inpainting and super-resolution on target images of size $256^2$. For both the LPIPS and SSIM, the first column indicates the number of test images (out of a total of 99) the corresponding method performed the best in terms of the respective metric, whereas the second column contains the average value over all the test images. The best performances are highlighted in boldface.\vspace{-0.1in}}}
    \label{table_256}
\end{table}


\begin{table}[t]
        \begin{tabular}{|l|l|l|l|l|l|l|}
            \cline{2-7}
            \multicolumn{1}{c|}{} & \multicolumn{3}{c|}{\textbf{Inpainting}} &  \multicolumn{3}{c|}{\textbf{Super-resolution}} \\ \cline{1-7} 
            \textbf{Method} & \textbf{Mask} & \multicolumn{2}{c|}{\textbf{LPIPS}} & \textbf{Input} & \multicolumn{2}{c|}{\textbf{LPIPS}} \\ \cline{1-7} 
            BRGM & \multirow{2}{*}{Half} & 17 & 0.495 & \multirow{2}{*}{$64^2$} & 32 & 0.426 \\ \cline{1-1} \cline{3-4} \cline{6-7}
            L-BRGM & & \b{82} & \b{0.460} & & \b{67} & \b{0.414} \\ \cline{1-7}
            \hline
            BRGM & \multirow{2}{*}{Eye-patch} & 10 & 0.428 & \multirow{2}{*}{$32^2$} & 33 & 0.470 \\ \cline{1-1} \cline{3-4} \cline{6-7}
            L-BRGM & & \b{89} & \b{0.393} & & \b{66} & \b{0.445} \\ \cline{1-7}
            \hline            
    \end{tabular}
    \caption{\small{L-BRGM vs. BRGM on images of resolution $1024^2$. L-BRGM outperforms BRGM on significantly more test images and results in superior LPIPS distance on average.\vspace{-0.1in}}}
    \label{table_2014}
\end{table}
\section{Numerical results}
\label{sec_results_numerics}
We consider two prototypical inverse problems, namely (i) image inpainting and (ii) super-resolution to validate L-BRGM quantitatively and compare it with state-of-the-art GAN-based image reconstruction approaches. For inpainting, we choose BRGM \cite{BRGM} and deep generative inpainting network (DeepGIN) \cite{li2020deepgin} as the competing methods. BRGM is a natural candidate for comparison since L-BRGM is a direct extension of it, while the choice of DeepGIN, which was an entry in the Aim-2020 inpainting challenge \cite{AIM2020}, was motivated by its applicability to inpainting with extreme masking.  
    
\noindent For super-resolution, we choose BRGM and the generative facial prior (GFP) approach \cite{GFPGAN} that seeks to restore facial images by exploiting the prior encapsulated by a GAN trained on face images (referred to as GFP-GAN) for comparing with L-BRGM. GFP-GAN uses the existing architecture from Real-ESRGAN \cite{RFB-ESRGAN} and is reported to yield state-of-the-art performance for facial image restoration. Contrary to the generators in BRGM and L-BRGM, both GFP-GAN and DeepGIN were trained on images of resolutions lower than $1024^2$. Consequently, we include comparisons with GFP-GAN and DeepGIN for target ground-truth images of resolution $256^2$, whereas for the images of resolution $1024^2$, the comparison is restricted to only BRGM. Both L-BRGM and BRGM reconstructions are computed with 2000 iterations. The Adam parameters are chosen as $\eta,\beta_,\beta_2=0.1,0.96,0.9999$; whereas the regularization penalties are: $\lambda_{\text{pix}},\lambda_{\text{vgg}},\lambda_{\text{map}},\lambda_{\text{lat}}=2\times 10^{-5}, 2 \times 10^7, 30, 0.4$. A set of 99 images (not used during training) are utilized for performance validation. The LPIPS scores are calculated using AlexNet features \cite{AlexNet} for a fair evaluation, since the perceptual loss based on VGG-16 embedding is already a part of the reconstruction objective $J$.   

For inpainting on images of size $256^2$, L-BRGM outperforms both BRGM and DeepGIN in terms of LPIPS (c.f. Table \ref{table_256}). In case of super-resolution, L-BRGM defeats BRGM, but performs slightly worse as compared with GFP-GAN. The superiority of GFP-GAN can presumably be attributed to the fact it is trained with bespoke loss terms that are particularly designed for face images (such as the facial component and identity preserving losses \cite{GFPGAN}). For inpainting, L-BRGM also emerges as the best performing method in terms of LPIPS on a majority of the test images. The inpainting examples in Fig. \ref{halfmasking} demonstrate that L-BRGM is conspicuously superior to its competitors. However, L-BRGM turns out to be slightly inferior in terms of SSIM, but we found that, unlike LPIPS, SSIM does not correlate well with the visual image quality and we therefore report only the LPIPS distances for inpainting and super-resolution on $1024^2$ images (see Table \ref{table_2014}). In this case, L-BRGM was found to significantly surpass BRGM for both inpainting masks and for the task of super-resolution. The superiority of L-BRGM over BRGM underscores the need for a better regularizer on the latent space, and highlights the advantages of data-driven modeling of such a regularizer.

\section{Conclusions and outlook}
We proposed a novel optimization-based approach for StyleGAN2 inversion, with potential applications to imaging inverse problems. The proposed L-BRGM algorithm leverages the full expressive power of the StyleGAN2 generator and mapping network for modeling the joint latent- and style-code prior in a Bayesian estimation framework. The proposed approach yields competitive performance with state-of-the-art GAN-based approaches for prototypical imaging inverse problems, such as inpainting and super-resolution. Nevertheless, developing a generic method to outperform the state-of-the-art on all possible target images is an ambitious and rather challenging objective. Therefore, further research is needed to not only produce novel regularization schemes that effectively make use of a pre-trained GAN, but also to learn generative models with more interpretable (and hence regularizable) latent spaces for image reconstruction problems.



\bibliographystyle{IEEEbib}
\bibliography{refs1}

\begin{thebibliography}{10}

\bibitem{scherzer2009variational}
Otmar Scherzer, Markus Grasmair, Harald Grossauer, Markus Haltmeier, and Frank
  Lenzen,
\newblock {\em Variational methods in imaging},
\newblock Springer, 2009.

\bibitem{nett_paper}
Housen Li, Johannes Schwab, Stephan Antholzer, and Markus Haltmeier,
\newblock ``{NETT}: solving inverse problems with deep neural networks,''
\newblock {\em Inverse Problems}, vol. 36, no. 6, 2020.

\bibitem{ar_nips}
Sebastian Lunz, Ozan {\"O}ktem, and Carola-Bibiane Sch{\"o}nlieb,
\newblock ``Adversarial regularizers in inverse problems,''
\newblock in {\em Advances in Neural Information Processing Systems}, 2018, pp.
  8507--8516.

\bibitem{bora_gancs}
A.~Bora, A.~Jalal, E.~Price, and A.~G. Dimakis,
\newblock ``Compressed sensing using generative models,''
\newblock in {\em International Conference on Machine Learning}, 2017.

\bibitem{red_schniter}
E.~T. Reehorst and P.~Schniter,
\newblock ``Regularization by denoising: clarifications and new
  interpretations,''
\newblock {\em IEEE Transactions on Computational Imaging}, vol. 5, no. 1, pp.
  52--67, 2019.

\bibitem{rare_deep_prior}
J.~Liu, Y.~Sun, C.~Eldeniz, W.~Gan, An~H., and Kamilov~U. S.,
\newblock ``Rare: image reconstruction using deep priors learned without
  groundtruth,''
\newblock {\em IEEE J. Selected Topics in Signal Processing}, vol. 14, no. 6,
  pp. 1088--1099, 2020.

\bibitem{GANInversion}
Weihao Xia, Yulun Zhang, Yujiu Yang, Jing-Hao Xue, Bolei Zhou, and Ming-Hsuan
  Yang,
\newblock ``Gan inversion: A survey,'' 2021.

\bibitem{BRGM}
Razvan~V Marinescu, Daniel Moyer, and Polina Golland,
\newblock ``Bayesian image reconstruction using deep generative models,'' 2021.

\bibitem{StyleGAN2}
Tero Karras, Miika Aittala, Janne Hellsten, Samuli Laine, Jaakko Lehtinen, and
  Timo Aila,
\newblock ``Training generative adversarial networks with limited data,''
\newblock in {\em Proc. NeurIPS}, 2020.

\bibitem{StyleGAN1}
Tero Karras, Samuli Laine, and Timo Aila,
\newblock ``A style-based generator architecture for generative adversarial
  networks,''
\newblock in {\em 2019 IEEE/CVF Conference on Computer Vision and Pattern
  Recognition (CVPR)}, 2019, pp. 4396--4405.

\bibitem{kingma2017adam}
Diederik~P. Kingma and Jimmy Ba,
\newblock ``Adam: A method for stochastic optimization,'' 2017.

\bibitem{LPIPS}
Richard Zhang, Phillip Isola, Alexei~A Efros, Eli Shechtman, and Oliver Wang,
\newblock ``The unreasonable effectiveness of deep features as a perceptual
  metric,''
\newblock in {\em CVPR}, 2018.

\bibitem{GFPGAN}
Xintao Wang, Yu~Li, Honglun Zhang, and Ying Shan,
\newblock ``Towards real-world blind face restoration with generative facial
  prior,''
\newblock in {\em The IEEE Conference on Computer Vision and Pattern
  Recognition (CVPR)}, 2021.

\bibitem{li2020deepgin}
Chu-Tak Li, Wan-Chi Siu, Zhi-Song Liu, Li-Wen Wang, and Daniel Pak-Kong Lun,
\newblock ``Deepgin: Deep generative inpainting network for extreme image
  inpainting,'' 2020.

\bibitem{AIM2020}
Evangelos Ntavelis, Andrés Romero, Siavash Bigdeli, and Radu Timofte,
\newblock ``Aim 2020 challenge on image extreme inpainting,'' 2020.

\bibitem{RFB-ESRGAN}
Taizhang Shang, Qiuju Dai, Shengchen Zhu, Tong Yang, and Yandong Guo,
\newblock ``Perceptual extreme super resolution network with receptive field
  block,'' 2020.

\bibitem{AlexNet}
Alex Krizhevsky, Ilya Sutskever, and Geoffrey~E Hinton,
\newblock ``Imagenet classification with deep convolutional neural networks,''
\newblock in {\em Advances in Neural Information Processing Systems}, 2012,
  vol.~25.

\end{thebibliography}

\end{document}